\def\year{2019}\relax
\documentclass[letterpaper]{article} 
\usepackage{aaai19}  
\usepackage{times}  
\usepackage{helvet}  
\usepackage{courier}  
\usepackage{url}  
\usepackage{graphicx}  
\usepackage{amsmath}
\usepackage{amsfonts}
\begin{document}
%
\title{End-to-end Structure-Aware Convolutional Networks \\for Knowledge Base Completion}

\def\wg{\textsuperscript{1}}
\def\ws{\textsuperscript{2}}

\author{Chao Shang,\wg \thanks{Work done during an internship at JD AI Research.}  Yun Tang,\ws \enspace  Jing Huang,\ws \enspace  Jinbo Bi,\wg \enspace  Xiaodong He,\ws \enspace  Bowen Zhou\ws\\
\wg Department of Computer Science and Engineering, University of Connecticut, Storrs, CT, USA\\
\ws JD AI Research, Mountain View, CA, USA\\
\{chao.shang, jinbo.bi\}@uconn.edu, \{yun.tang, jing.huang, xiaodong.he, bowen.zhou\}@jd.com
}
\maketitle
\begin{abstract}

Knowledge graph embedding has been an active research topic for knowledge base completion, with progressive improvement from the initial {\it TransE, TransH, DistMult} et al to the current state-of-the-art {\it ConvE}. {\it ConvE} uses 2D convolution over embeddings and multiple layers of nonlinear features to model knowledge graphs. The model can be efficiently trained and scalable to large knowledge graphs. However, there is no structure enforcement in the embedding space of {\it ConvE}. The recent graph convolutional network ({\it GCN}) provides another way of learning graph node embedding by successfully utilizing graph connectivity structure. In this work, we propose a novel end-to-end Structure-Aware Convolutional Network ({\it SACN}) that takes the benefit of {\it GCN} and {\it ConvE} together. {\it SACN} consists of an encoder of a weighted graph convolutional network ({\it WGCN}), and a decoder of a convolutional network called {\it Conv-TransE}. {\it WGCN} utilizes knowledge graph node structure, node attributes and edge relation types. It has learnable weights that adapt the amount of information from neighbors used in local aggregation, leading to more accurate embeddings of graph nodes. Node attributes in the graph are represented as additional nodes in the {\it WGCN}. The decoder {\it Conv-TransE} enables the state-of-the-art {\it ConvE} to be translational between entities and relations while keeps the same link prediction performance as {\it ConvE}. We demonstrate the effectiveness of the proposed {\it SACN} on standard FB15k-237 and WN18RR datasets, and it gives about 10\% relative improvement over the state-of-the-art {\it ConvE} in terms of HITS@1, HITS@3 and HITS@10.

\end{abstract}

\section{Introduction}
\noindent

Over the recent years, large-scale knowledge bases (KBs), such as Freebase \cite{bollacker2008freebase}, DBpedia \cite{auer2007dbpedia}, NELL \cite{carlson2010toward} and YAGO3 \cite{mahdisoltani2013yago3}, have been built to store structured information about common facts. KBs are multi-relational graphs whose nodes represent entities and edges represent relationships between entities, and the edges are labeled with different relations. The relationships are organized in the forms of $(s,r,o)$ triplets (e.g. entity $s$ = Abraham Lincoln, relation $r$ = DateOfBirth, entity $o$ = 02-12-1809). These KBs are extensively used for web search, recommendation and question answering. Although these KBs have already contained millions of entities and triplets, they are far from complete compared to existing facts and newly added knowledge of the real world. Therefore knowledge base completion is important in order to predict new triplets based on existing ones and thus further expand KBs.

One of the recent active research areas for knowledge base completion is knowledge graph embedding: it encodes the semantics of entities and relations in a continuous low-dimensional vector space (called embeddings). These embeddings are then used for predicting new relations. Started from a simple and effective approach called
{\it TransE} \cite{bordes2013translating}, many knowledge graph embedding methods have been proposed, such as {\it TransH} \cite{wang2014knowledge}, {\it TransR} \cite{lin2015learning}, {\it DistMult} \cite{yang2014distmult}, {\it TransD} \cite{ji2015knowledge}, {\it ComplEx} \cite{trouillon2016complex}, {\it STransE} \cite{nguyen2016stranse}. Some surveys \cite{nguyen2017overview,wang2017knowledge} give details and comparisons of these embedding methods.

The most recent {\it ConvE} \cite{dettmers2017conve} model uses 2D convolution over embeddings and multiple layers of nonlinear features, and achieves the state-of-the-art performance on common benchmark datasets for knowledge graph link prediction. In {\it ConvE}, the embeddings of $s$ and $r$ are reshaped and concatenated into an input matrix and fed to the convolution layer. Convolutional filters of $n \times n$  are used to output feature maps that are across different dimensional embedding entries. Thus {\it ConvE} does not keep the translational property as {\it TransE} which is an additive embedding vector operation: $e_{s} + e_{r} \approx e_{o}$ (\cite{nguyen2017novel}). In this paper, we remove the reshape step of {\it ConvE} and operate convolutional filters directly in the same dimensions of $s$ and $r$. This modification gives better performance compared with the original {\it ConvE}, and has an intuitive interpretation which keeps the global learning metric the same for $s$, $r$, and $o$ in an embedding triple $(e_{s},e_{r},e_{o})$. We name this embedding as {\it Conv-TransE}.

{\it ConvE} also does not incorporate connectivity structure in the knowledge graph into the embedding space. In contrast, graph convolutional network (GCN) has been an effective tool to create node embeddings which aggregate local information in the graph neighborhood for each node  \cite{kipf2016semi,hamilton2017inductive,kipf2016nipsws,pham2017aaai,shang2018edge}.
GCN models have additional benefits \cite{hamilton2017representation}, such as leveraging the attributes associated with nodes. They can also impose the same aggregation scheme when computing the convolution for each node, which can be considered a method of regularization, and improves efficiency. Although scalability is originally an issue for GCN models, the latest data-efficient GCN, PinSage \cite{ying2018kdd}, is able to handle billions of nodes and edges.

In this paper, we propose an end-to-end graph Structure-Aware Convolutional Networks ({\it SACN}) that take all benefits of GCN and {\it ConvE} together. {\it SACN} consists of an encoder of a weighted graph convolutional network ({\it WGCN}), and a decoder of a convolutional network called {\it Conv-TransE}. {\it WGCN} utilizes knowledge graph node structure, node attributes and relation types. It has learnable weights to determine the amount of information from neighbors used in local aggregation, leading to more accurate embeddings of graph nodes. Node attributes are added to {\it WGCN} as additional for easy integration. The output of {\it WGCN} becomes the input of the decoder {\it Conv-TransE}. {\it Conv-TransE} is similar to {\it ConvE} but with the difference that {\it Conv-TransE} keeps the translational characteristic between entities and relations. We show that {\it Conv-TransE} performs better than {\it ConvE}, and our {\it SACN} improves further on top of {\it Conv-TransE} in the standard benchmark datasets. 
The code for our model and experiments is publicly available \footnote{https://github.com/JD-AI-Research-Silicon-Valley/SACN}.

Our contributions are summarized as follows:
\begin{itemize}
\item We present an end-to-end network learning framework {\it SACN} that takes benefit of both GCN and {\it Conv-TransE}. The encoder GCN model leverages graph structure and attributes of graph nodes. The decoder \textit{Conv-TransE} simplifies {\it ConvE} with special convolutions and keeps the translational property of {\it TransE} and the prediction performance of {\it ConvE};
\item We demonstrate the effectiveness of our proposed {\it SACN} on the standard FB15k-237 and WN18RR datasets, and show about 10\% relative improvement over the state-of-the-art {\it ConvE} in terms of HITS@1, HITS@3 and HITS@10.
\end{itemize}

\section{Related Work}

Knowledge graph embedding learning has been an active research area with applications directly in knowledge base completion (i.e. link prediction) and relation extractions. TransE \cite{bordes2013translating} started this line of work by projecting both entities and relations into the same embedding vector space, with translational constraint of $e_{s} + e_{r} \approx e_{o}$. Later works enhanced KG embedding models such as TransH \cite{wang2014knowledge}, TransR \cite{lin2015learning}, and TransD \cite{ji2015knowledge} introduced new representations of relational translation and thus increased model complexity. These models were categorized as {\it translational distance}  models \cite{wang2017knowledge} or {\it additive} models, while DistMult \cite{yang2014distmult} and ComplEx \cite{trouillon2016complex} are {\it multiplicative} models \cite{sharma2018towards}, due to the multiplicative score functions used for computing entity-relation-entity triplet likelihood.

The most recent KG embedding models are {\it ConvE} \cite{dettmers2017conve} and {\it ConvKB} \cite{nguyen2017novel}. {\it ConvE} was the first model using 2D convolutions over embeddings of different embedding dimensions, with the hope of extracting more feature interactions. {\it ConvKB} replaced 2D convolutions in {\it ConvE} with 1D convolutions, which constrains the convolutions to be the same embedding dimensions and keeps the translational property of TransE. {\it ConvKB} can be considered as a special case of {\it Conv-TransE} that only uses filters with width equal to $1$. Although {\it ConvKB} was shown to be better than {\it ConvE}, the results on two datasets (FB15k-237 and WN18RR) were not consistent, so we leave these results out of our comparison table. The other major difference of {\it ConvE} and {\it ConvKB} is on the loss functions used in the models. {\it ConvE} used the cross-entropy loss that could be sped up with 1-N scoring in the decoder, while {\it ConvKB} used a hinge loss that was computed from positive examples and sampled negative examples. We take the decoder from {\it ConvE} because we can easily integrate the encoder of GCN and the decoder of {\it ConvE} into an end-to-end training framework, while {\it ConvKB} is not suitable for our approach.


These embedding models achieved good performance for knowledge base completion in terms of efficiency and scalability. However, these approaches only modeled relational triplets, while ignoring a large number of attributes associated with graph nodes, e.g., ages of people or release region of music.
Furthermore, these models do not enforce any large-scale connectivity structure in the embedding space, and totally ignore the knowledge graph structure. The proposed ({\it SACN}) handles these two problems in an end-to-end training framework, by using a variant of graph convolutional network (GCN) as the encoder, and a variant of {\it ConvE} as the decoder.

\begin{figure*}
    \includegraphics[width=\textwidth]{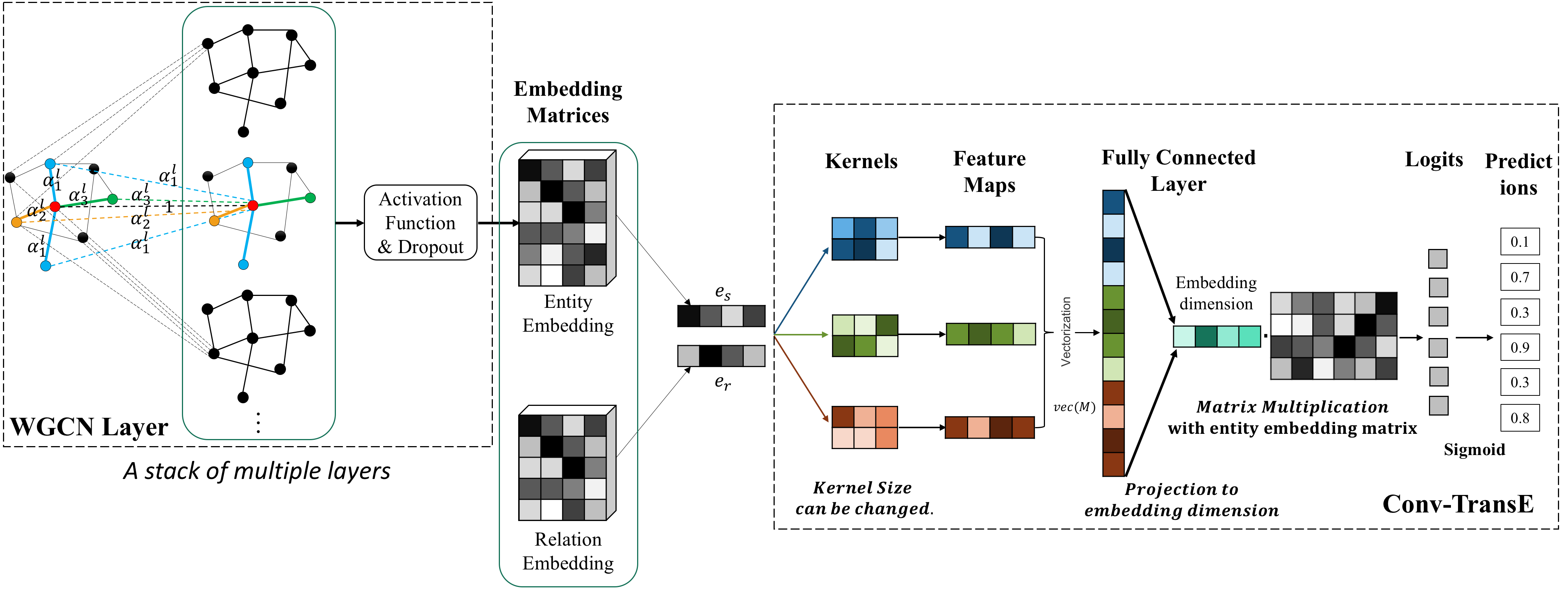}
    \caption{An illustration of our end-to-end Structure-Aware Convolutional Networks model. For encoder, a stack of multiple WGCN layers builds an entity/node embedding matrix. For decoder, $e_s$ and $e_r$ are fed into {\it Conv-TransE}. The output embeddings are vectorized and projected, and matched with all candidate $e_o$ embeddings via inner products. A logistic sigmoid function is used to get the scores.
    }
    \medskip
    \label{fig:structure}
\end{figure*}

GCNs were first proposed in \cite{bruna2013spectral} where graph convolutional operations were defined in the Fourier domain. The eigendecomposition of the graph Laplacian caused intense computation. Later, smooth parametric spectral filters \cite{henaff2015deep,defferrard2016convolutional} were introduced to achieve localization in the spatial domain and improve computational efficiency.
Recently, Kipf et al. \cite{kipf2016semi} simplified these spectral methods by a first-order approximation with the Chebyshev polynomials. 
The spatial graph convolution approaches \cite{hamilton2017inductive} define convolutions directly on graph, which sum up node features over all spatial neighbors using adjacency matrix.

GCN models were mostly criticized for its huge memory requirement to scale to massive graphs. However, \cite{ying2018kdd} developed a data efficient GCN algorithm called PinSage, which combined efficient random walks and graph convolutions to generate embeddings of nodes that incorporated both graph structure as well as node features. The experiments on Pinterest data were the largest application of deep graph embeddings to date with 3 billion nodes and 18 billion edges \cite{ying2018kdd}. This success paves the way for a new generation of web-scale recommender systems based on GCNs. Therefore we believe that our proposed model could take advantage of huge graph structures and high computational efficiency of {\it Conv-TransE}.


\section{Method}

In this section, we describe the proposed end-to-end {\it SACN}. The encoder {\it WGCN} is focused on representing entities by aggregating connected entities as specified by the relations in the KB. With node embeddings as the input, the decoder {\it Conv-TransE} network aims to represent the relations more accurately by recovering the original triplets in the KB. Both encoder and decoder are trained jointly by minimizing the discrepancy (cross-entropy) between the embeddings $e_s +e_r$ and $e_o$ to preserve the translational property $e_s +e_r \approx e_o$. We consider an undirected graph $G = (V, E)$ throughout this section, where $V$ is a set of nodes with $|V| = N$, and $E \subseteq V \times V$ is a set of edges with $|E| = M$. 

\subsection{Weighted Graph Convolutional Layer}
The {\it WGCN} is an extension of classic GCN \cite{kipf2016semi} in the way that it weighs the different types of relations differently when aggregating and the weights are adaptively learned during the training of the network. By this adaptation, the {\it WGCN} can control the amount of information from neighboring nodes used in aggregation. Roughly speaking, the {\it WGCN} treats a multi-relational KB graph as multiple single-relational subgraphs where each subgraph entails a specific type of relations. The {\it WGCN} determines how much weights to give to each subgraph when combining the GCN embeddings for a node. 

The $l$-th {\it WGCN} layer takes the output vector of length $F^{l}$ for each node from the previous layer as inputs and generates a new representation comprising $F^{l+1}$ elements. Let $h^{l}_i$ represent the input (row) vector of the node $v_i$ in the $l$-th layer, and thus $H^{l}\in {\mathbb{R}}^{N \times F^{l}}$ be the input matrix for this layer. The initial embedding $H^1$ is randomly drawn from Gaussian. If there are a total of $L$ layers in the {\it WGCN}, the output $H^{L+1}$ of the $L$-th layer is the final embedding.
Let the total number of edge types be $T$ in a multi-relational KB graph with $E$ edges.
The interaction strength between two adjacent nodes is determined by their relation type and this strength is specified by a parameter $\{\alpha_t, 1 \leq t \leq T\} $ for each edge type, which is automatically learned in the neural network. 

Figure \ref{fig:structure} illustrates the entire process of {\it SACN}. In this example, the {\it WGCN} layers of the network compute the embeddings for the red node in the middle graph. These layers aggregate the embeddings of neighboring entity nodes as specified in the KB relations. Three colors (blue, yellow and green) of the edges indicate three different relation types in the graph. The corresponding three entity nodes are summed up with different weights according to $\alpha_t$ in this layer to obtain the embedding of the red node. The edges with the same color (same relation type) use the same $\alpha_t$. Each layer has its own set of relation weights $\alpha_{t}^l$. Hence, the output of the $l$-th layer for the node $v_i$  can be written as follows:
\begin{equation} \label{equ_sum}
h^{l+1}_i = \sigma\left(\sum_{j \in \mathbf{N_i}} \alpha_{t}^l g( h^{l}_i,  h^{l}_j ) \right) ,
\end{equation}
where $h^{l}_j \in \mathbb{R}^{F^{l}}$ is the input for node $v_j$, and $v_j$ is a node in the neighbor $N_i$ of node $v_i$. The $g$ function specifies how to incorporate neighboring information. Note that the activation function $\sigma$ here is applied to every component of its input vector. 
Although any function $g$ suitable for a KB embedding can be used in conjunction with the proposed framework, we implement the following $g$ function:
\begin{equation} \label{equ_gm}
g(h^{l}_i,  h^{l}_j ) = h_j^l W^{l} ,
\end{equation}
where $W^l \in \mathbb{R}^{F^l \times F^{l+1}}$ is the connection coefficient matrix and used to linearly transform $h^l_i$ to $h^{l+1}_i \in \mathbb{R}^{F^{l+1}}$.

In Eq. \eqref{equ_sum}, the input vectors of all neighboring nodes are summed up but not the node $v_i$ itself, hence self-loops are enforced in the network. For node $v_i$, the propagation process is defined as:
\begin{equation} 
h^{l+1}_i = \sigma(\sum_{j \in \mathbf{N_i}} \alpha^l_{t} h_j^l W^{l} + h_i^l W^{l}) .
\end{equation}
The output of the layer $l$ is a node feature matrix: $ H^{l+1} \in {\mathbb{R}}^{N \times F^{l+1}}$, and $h^{l+1}_i$  is the $i$-th row of $H^{l+1}$, which represents features of the node $v_i$ in the $(l+1)$-th layer.

\begin{figure}
    \includegraphics[width=0.48\textwidth]{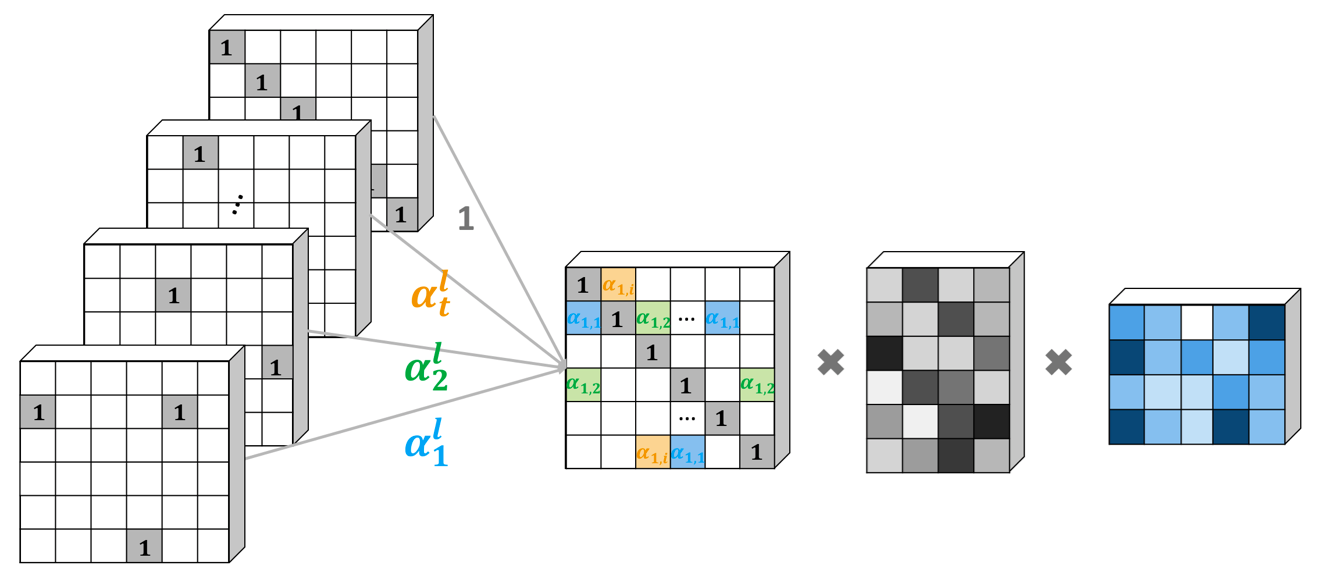}
    \caption{A weighted graph convolutional network ({\it WGCN}) for entity embedding.}
    \medskip
    \label{fig:wgcn-mat}
\end{figure}

The above process can be organized as a matrix multiplication as shown in Figure \ref{fig:wgcn-mat} to simultaneously compute embeddings for all nodes through an adjacency matrix. For each relation (edge) type, an adjacency matrix  $A_t$ is a binary matrix whose $ij$-th entry is 1 if an edge connecting $v_i$ and $v_j$ exists or 0 otherwise. The final adjacency matrix is written as follows: 
\begin{equation} 
A^l = \sum^{T}_{t=1} ( \alpha_t^l A_{t} ) + I ,
\end{equation}
where $I$ is the identity matrix of size $N \times N$. Basically, the $A^l$ is the weighted sum of the adjacency matrices of subgraphs plus self-connections. In our implementation, we consider all first-order neighbors in the linear transformation for each layer as shown in Figure \ref{fig:wgcn-mat}:
\begin{equation} \label{conv_layer}
H^{l+1} = \sigma(A^l H^l W^l).
\end{equation}

\subsubsection{Node Attributes.}
In a KB graph, nodes are often associated with several attributes in the form of $(entity, relation, attribute)$. For example, ($s=$ Tom, $r=$ people.person.gender, $a=$ male) is an instance where gender is an attribute associated with a person. If a vector representation is used for node attributes, there would be two potential problems. First, the number of attributes for each node is usually small, and differs from one to another. Hence, the attribute vector would be very sparse. Second, the value of zero in the attribute vectors may have ambiguous meanings: the node does not have the specific attribute, or the node misses the value for this attribute. These zeros would affect the accuracy of the embedding. 

In this work, the entity attributes in the knowledge graph are represented by another set of nodes in the network called attribute nodes. Attribute nodes act as the ``bridges" to link the related entities. The entity embeddings can be transported over these ``bridges" to incorporate the entity's attribute into its embedding. Because these attributes exhibit in triplets, we represent the attributes similarly to the representation of the entity $o$ in relation triplets. Note that each type of attribute corresponds to a node. For instance, in our example, gender is represented by a single node rather than two nodes for ``male" and ``female".
In this way, the {\it WGCN} not only utilizes the graph connectivity structure (relations and relation types), but also leverages the node attributes (a kind of graph structure) effectively. That is why we name our {\it WGCN} as a structure-aware convolution network.

\subsection{Conv-TransE}
We develop the {\it Conv-TransE} model as a decoder that is based on {\it ConvE} but with the translational property of {\it TransE}: $e_s + e_r  \approx e_o$. 
The key difference of our approach from {\it ConvE} is that there is no reshaping after stacking $e_s$ and $e_r$.
Filters (or kernels) of size $2 \times k$ , $k \in \{1,2,3,...\}$, are used in the convolution. 
The example in Figure \ref{fig:structure} uses $2 \times 3$ kernels to compute 2D convolutions.
We experimented with several of such settings in our empirical study.

Note that in the encoder of {\it SACN}, 
the dimension of the relation embedding is commonly chosen to be the same as the dimension of the entity embedding, so in other words, is equal to $F^L$. Hence, the two embeddings can be stacked. For the decoder, the inputs are two embedding matrices: one ${\mathbb{R}}^{N \times F^{L}}$ from {\it WGCN} for all entity nodes, and the other ${\mathbb{R}}^{M \times F^{L}}$ for
relation embedding matrix which is trained as well.
Because we use a mini-batch stochastic training algorithm, the first step of the decoder performs a look-up operation upon the embedding matrices to retrieve the input $e_s$ and $e_r$ for the triplets in the mini-batch.

More precisely, given $C$ different kernels where the $c$-th kernel is parameterized by $\omega_c$, the convolution in the decoder is computed as follows:
\begin{equation} 
\label{equ_filter}
\begin{split}
m_c (e_s, e_r, n)  = &  \sum_{\tau=0}^{K-1} \omega_{c}(\tau, 0)\hat{e}_s(n+\tau) \\
& +  \omega_{c}( \tau, 1) \hat{e}_r(n+\tau),  
\end{split}
\end{equation}
where $K$ is the kernel width, $n$ indexes the entries in the output vector and $n \in [0,F^L-1]$, and the kernel parameters $\omega_c$ are trainable. $\hat{e}_s$ and $\hat{e}_r$ 
are padding version of $e_s$ and $e_r$ respectively. 
If the dimension $s$ of kernel is odd, the first $\lfloor K/2 \rfloor$ and last $\lfloor K/2 \rfloor$ components are filled with 0. Here $\lfloor value \rfloor$ returns the floor of $value$.
Otherwise, the first $\lfloor K/2 \rfloor -1$ and last $\lfloor K/2 \rfloor$ components are filled with 0. 
Other components are copied from $e_s$ and $e_r$ directly.
As shown in Eq. \eqref{equ_filter} the convolution operation amounts to a sum of $e_s$ and $e_r$ after the one-dimensional convolution. Hence, it preserves the translational property of the embeddings of $e_s$, $e_r$. The output forms a vector $M_c(e_s,e_r) = [m_c(e_s,e_r,0), ..., m_c(e_s,e_r,F^L-1)]$. Aligning the output vectors from the convolution with all kernels yield a matrix
$\mathbf{M}(e_s, e_r) \in {\mathbb{R}}^{C \times F^L}$.

\begin{table}[t]
	\caption{Scoring function $\psi(e_s, e_o)$. Here $\bar{e_s}$ and $\bar{e_r}$ denote a 2D reshaping of $e_s$ and $e_r$.}
	\label{tbl:scoring}
	\smallskip
	\begin{center}
		\begin{tabular}{|l|c|}
			\hline
			Model & Scoring Function $\psi(e_s, e_o)$\\
			\hline
			TransE 	 & $||e_s + e_r - e_o||_p$  \\
			DistMult 	 & $<e_s, e_r, e_o>$  \\
			ComplEx  & $<e_s, e_r, e_o>$  \\
			ConvE 	 &  $  f( vec( f(concat(\bar{e_s}, \bar{e_r}) * \omega )) W )e_o  $   \\
			{\it ConvKB}  & $concat(g([e_s, e_r, e_o]* \omega)) \beta$ \\
			SACN  & $f(vec(\mathbf{M}(e_s, e_r)) W )e_o$ \\
			\hline
		\end{tabular}
	\end{center}
	\vskip -0.25in
\end{table}

Finally, the scoring function for the {\it Conv-TransE} method after the nonlinear convolution is defined as below:
\begin{equation} 
\psi(e_s, e_o) = f(vec(\mathbf{M}(e_s, e_r)) W )e_o,
\end{equation}
where $W \in \mathbb{R}^{CF^{L} \times F^{L}}$ is a matrix for the linear transformation, and $f$ denotes a non-linear function. The feature map matrix is reshaped into a vector $vec(\mathbf{M}) \in \mathbb{R}^{C F^{L}}$ and projected into a $F^{L}$ dimensional space using $W$ for linear transformation. Then the calculated embedding is matched to $e_o$ by an appropriate distance metric. During the training in our experiments, we apply the logistic sigmoid function to the scoring:
\begin{equation} 
p(e_s, e_r, e_o) = \sigma(\psi(e_s, e_o)).
\end{equation}

In Table \ref{tbl:scoring}, we summarize the scoring functions used by several state of the art models. The vector $e_s$ and $e_o$ are the subject and object embedding respectively, $e_r$ is the relation embedding, ``concat" means concatenates the inputs, and ``*" denotes the convolution operator.

In summary, the proposed {\it SACN} model takes advantage of knowledge graph node connectivity, node attributes and relation types. The learnable weights in {\it WGCN} help to collect adaptive amount of information from neighboring graph nodes. The entity attributes are added as additional nodes in the network and are easily integrated into the {\it WGCN}. {\it Conv-TransE} keeps the translational property between entities and relations to learn node embeddings for the link prediction.
We also emphasize that our {\it SACN} has significant improvements over {\it ConvE} with or without the use of node attributes.

\section{Experiments}

\begin{table}[t]
	\caption{Statistics of datasets.}
	\label{tbl:sample}
	\bigskip
	\begin{center}
		\vskip -0.1in
		\setlength{\tabcolsep}{2pt}
		\begin{tabular}{|l|c|c|c|}
			\hline
			Dataset & FB15k-237 & WN18RR & FB15k-237-Attr\\
			\hline
			Entities	 & 14,541	& 40,943  & 14,744 \\
			Relations	 & 237 & 11 & 484 \\
			Train Edges	 & 272,115 & 86,835 & 350,449 \\
			Val. Edges	 & 17,535 & 3,034 & 17,535 \\
			Test Edges	 & 20,466 & 3,134 & 20,466 \\
			Attributes Triples & --- & --- & 78,334 \\
			Attributes & --- & --- & 203 \\
			\hline
		\end{tabular}
	\end{center}
	\vskip -0.25in
\end{table}

\subsection{Benchmark Datasets}
Three benchmark datasets (FB15k-237, WN18RR and FB15k-237-Attr) are utilized in this study to evaluate the performance of link prediction.\\
\\
\textbf{FB15k-237.}
The FB15k-237 \cite{toutanova2015observed} dataset contains knowledge base relation triples and textual mentions of Freebase entity pairs, as used in the work published in \cite{toutanova2015observed}. The knowledge base triples are a subset of the FB15K \cite{bordes2013translating}, originally derived from Freebase. The inverse relations are removed in FB15k-237.\\
\textbf{WN18RR.}
WN18RR \cite{dettmers2017conve} is created from WN18 \cite{bordes2013translating}, which is a subset of WordNet. WN18 consists of 18 relations and 40,943 entities. However, many text triples obtained by inverting triples from the training set.
Thus WN18RR dataset \cite{dettmers2017conve} is created to ensure that the evaluation dataset does not have inverse relation test leakage. In summary, WN18RR dataset contains 93,003 triples with 40,943 entities and 11 relation types.
\begin{table*}
    \caption{Link prediction for FB15k-237, WN18RR and FB15k-237-Attr datasets.}
    \bigskip
    \label{tab:table3}
    \tabcolsep=0.32cm
    \centering
    \begin{tabular}{c|c|c|c|c|c|c|c|c}
        \hline
         & \multicolumn{4}{c|}{\textbf{FB15k-237}} & \multicolumn{4}{c}{\textbf{WN18RR}}\\
        \cline{2-9}
         & \multicolumn{3}{c|}{Hits} & \multicolumn{1}{c|}{}  & \multicolumn{3}{c|}{Hits}  & \multicolumn{1}{c}{} \\
        \cline{2-9}
        Model & @10 & @3 & @1 & MRR & @10 & @3 & @1 & MRR  \\
        \hline
        \hline
        DistMult \cite{yang2014distmult}  & 0.42 & 0.26 & 0.16& 0.24  & 0.49 & 0.44 & 0.39 & 0.43  \\
        \hline
        ComplEx \cite{trouillon2016complex}  & 0.43 & 0.28 & 0.16 & 0.25  & 0.51 & 0.46 & 0.41 & 0.44\\
        \hline
        R-GCN \cite{schlichtkrull2018modeling} & 0.42 & 0.26 & 0.15 & 0.25 & --- & ---  & --- & ---  \\
        \hline
        ConvE \cite{dettmers2017conve}  & 0.49 & 0.35 & 0.24 & 0.32   & 0.48 & 0.43 & 0.39 & 0.46 \\
        \hline
        \hline
        Conv-TransE & 0.51 & 0.37 & 0.24 & 0.33  & 0.52 & 0.47 & 0.43 & 0.46\\
        \hline
        SACN  & 0.54 & 0.39 & 0.26 & 0.35  & \textbf{0.54} & \textbf{0.48} & \textbf{0.43} & \textbf{0.47} \\
        \hline
        SACN using \textbf{FB15k-237-Attr} & \textbf{0.55} & \textbf{0.40} & \textbf{0.27} & \textbf{0.36} & --- &--- & --- & --- \\
        \hline
        \hline
        \textbf{Performance Improvement}  & \textbf{12.2\%} & \textbf{14.3\%} & \textbf{12.5\%}& \textbf{12.5\%}   & \textbf{12.5\%} & \textbf{11.6\%} & \textbf{10.3\%} & \textbf{2.2\%} \\
        \hline
    \end{tabular}
\end{table*}

\subsection{Data Construction}
Most of the previous methods only model the entities and relations, and ignore the abundant entity attributes. Our method can easily model a large number of entity attribute triples. In order to prove the efficiency, we extract the attribute triples from the FB24k \cite{lin2016knowledge} dataset to build the evaluation dataset called FB15k-237-Attr.\\
\\
\textbf{FB24k.} 
FB24k \cite{lin2016knowledge} is built based on Freebase dataset. FB24k only selects the entities and relations which constitute at least 30 triples. The number of entities is 23,634, and the number of relations is 673. In addition, the reversed relations are removed from the original dataset. In the FB24k datasets, the attribute triples are provided. FB24k contains 207,151 attribute triples and 314 attributes. \\
\textbf{FB15k-237-Attr.} 
We extract the attribute triples of entities in FB15k-237 from FB24k. During the mapping, there are 7,589 nodes from the original 14,541 entities which have the node attributes. Finally, we extract 78,334 attribute triples from FB24k. These triples include 203 attributes and 247 relations. 
Based on these triples, we create the ``FB15k-237-Attr" dataset, which includes 14,541 entity nodes, 203 attribute nodes, 484 relation types. All the 78,334 attribute triples are combined with the training set of FB15k-237.

\subsection{Experimental Setup}

The hyperparameters in our {\it Conv-TransE} and {\it SACN} models are determined by a grid search during the training. We manually specify the hyperparameter ranges: learning rate $\{0.01,0.005,0.003,0.001\}$, dropout rate $\{0.0,0.1,0.2,0.3,0.4,0.5\}$, embedding size $\{100, 200, 300\}$, number of kernels $\{50, 100, 200, 300\}$, and kernel size $\{2\times1, 2\times3, 2\times5\}$.

Here all the models use the {\it WGCN} with two layers.
For different datasets, we have found that the following  settings work well: for FB15k-237, set the dropout to 0.2, number of kernels to 100, learning rate to 0.003 and embedding size to 200 for {\it SACN};
for WN18RR dataset, set dropout to 0.2, number of kernels to 300, learning rate to 0.003, and embedding size to 200 for {\it SACN}. 
When using the {\it Conv-TransE}-alone model, these settings still work well.

Each dataset is split into three sets for: training, validation and testing, which is same with the setting of the original {\it ConvE}.
We use the adaptive moment (Adam) algorithm \cite{kingma2014adam} for training the model.  Our models are implemented by PyTorch and run on NVIDIA Tesla P40 Graphics Processing Units.
For the FB15k-237 dataset, the computation time of {\it SACN} for each epoch is about 1 minute. For the WN18RR, the computation time of {\it SACN} for one epoch is about 1.5 minutes.

\subsection{Results}
\subsubsection{Evaluation Protocol} 
Our experiments use the the proportion of correct entities ranked in top 1,3 and 10 (Hits@1, Hits@3, Hits@10) and the mean reciprocal rank (MRR) as the metrics. In addition, since some corrupted triples exist in the knowledge graphs, we use the filtered setting \cite{bordes2013translating}, i.e. we filter out all valid triples before ranking.

\subsubsection{Link Prediction}
Our results on the standard FB15k-237, WN18RR and FB15k-237-Attr are shown in Table \ref{tab:table3}.
Table \ref{tab:table3} reports Hits@10, Hits@3, Hits@1 and MRR results of four different baseline models and two our models on three knowledge graphs datasets. The FB15k-237-Attr dataset is used to prove the efficiency of node attributes. 
So we run our SACN in FB15k-237-Attr to do the comparison with SACN using FB15k-237.

We first compare our {\it Conv-TransE} model with the four baseline models. {\it ConvE} has the best performance comparing all baselines.  In FB15k-237 dataset, our {\it Conv-TransE} model improves upon ConvE's Hits@10 by a margin of 4.1\% , and upon ConvE's Hits@3 by a margin of 5.7\% for the test. In WN18RR dataset, {\it Conv-TransE} improves upon ConvE's Hits@10 by a margin of 8.3\% , and upon ConvE's Hits@3 by a margin of 9.3\% for the test. For these results, we conclude that {\it Conv-TransE} using neural network keeps the translational characteristic between entities and relations and achieve better performance.

Second, the structure information is added into our {\it SACN} model. In Table \ref{tab:table3}, {\it SACN} also get the best performances in the test dataset comparing all baseline methods. In FB15k-237, comparing {\it ConvE}, our {\it SACN} model improves Hits@10 value by a margin of 10.2\%, Hits@3 value by a margin of 11.4\%, Hits@1 value by a margin of 8.3\% and MRR value by a margin of 9.4\% for the test.
In WN18RR dataset, comparing {\it ConvE}, our {\it SACN} model improves Hits@10 value by a margin of 12.5\%, Hits@3 value by a margin of 11.6\%, Hits@1 value by a margin of 10.3\% and MRR value by a margin of 2.2\% for the test.
So our method has significant improvements over {\it ConvE} without attributes. 

Third, we add node attributes into our {\it SACN} model, i.e. we use the FB15k-237-Attr to train {\it SACN}. Note that {\it SACN} has significant improvements over ConvE without attributes. Adding attributes improves performance again. Our model using attributes improves upon ConvE's Hits@10 by a margin of 12.2\% , Hits@3 by a margin of 14.3\%, Hits@1 by a margin of 12.5\% and MRR by a margin of 12.5\%. 
In addition, our {\it SACN} using attributes improved Hits@10 by a margin of 1.9\% , Hits@3 by a margin of 2.6\%, Hits@1 by a margin of 3.8\% and MRR by a margin of 2.9\% comparing with {\it SACN} without attributes.

In order to better compare with {\it ConvE}, we also use the attributes into {\it ConvE}. Here the attributes will be treated as the entity triplets. Following the official {\it ConvE} code with default setting, the test result in FB15k-237-Attr was: 0.46 (Hits@10), 0.33 (Hits@3), 0.22 (Hits@1) and 0.30 (MRR).
Comparing to the performance without the attributes, adding the attributes into the {\it ConvE} didn't improve performance.

\begin{figure}[htb]
\centering
  \begin{tabular}{@{}cccc@{}}
    \includegraphics[width=.235\textwidth, height=3.55cm]{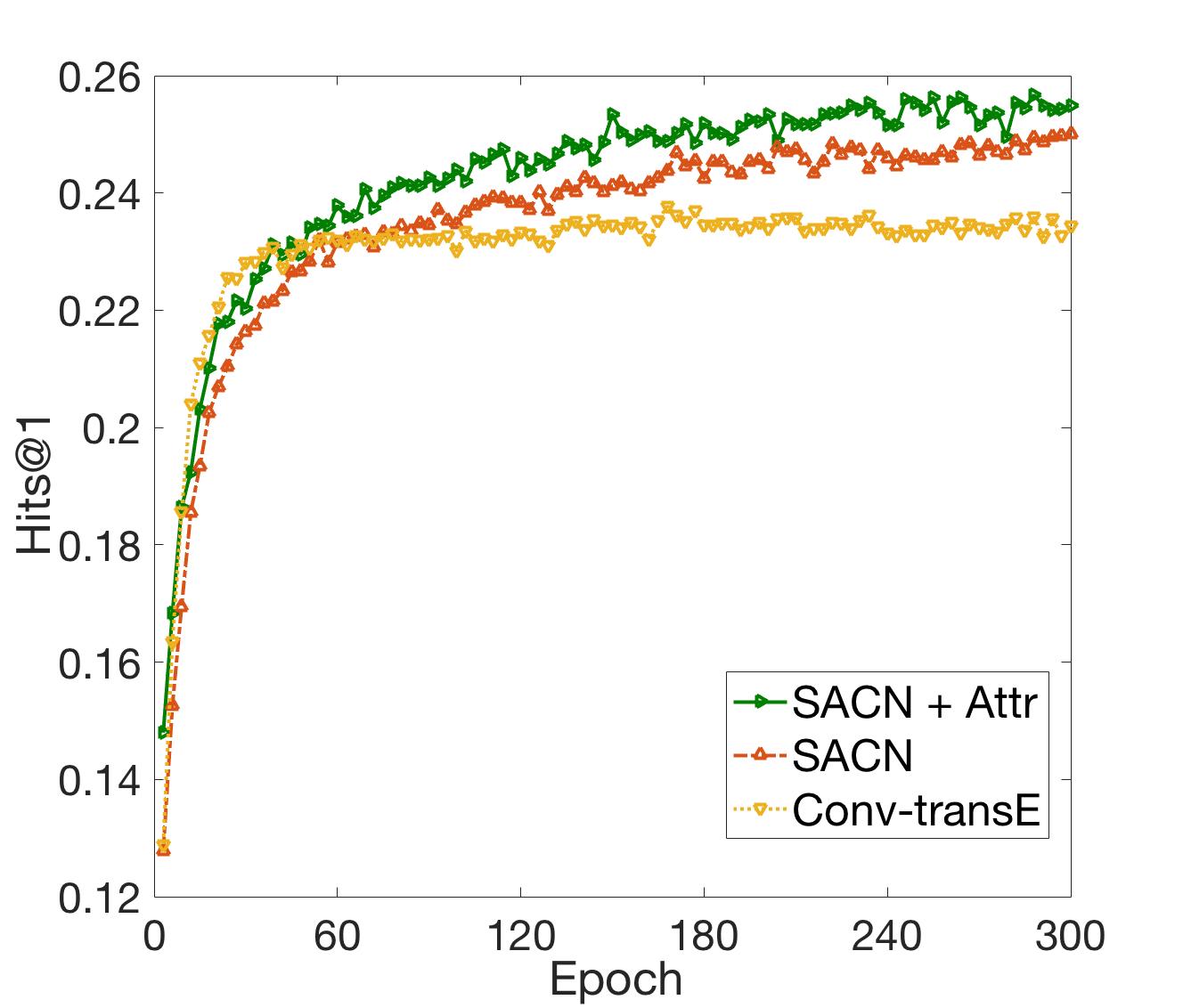} &
    \includegraphics[width=.235\textwidth, height=3.55cm]{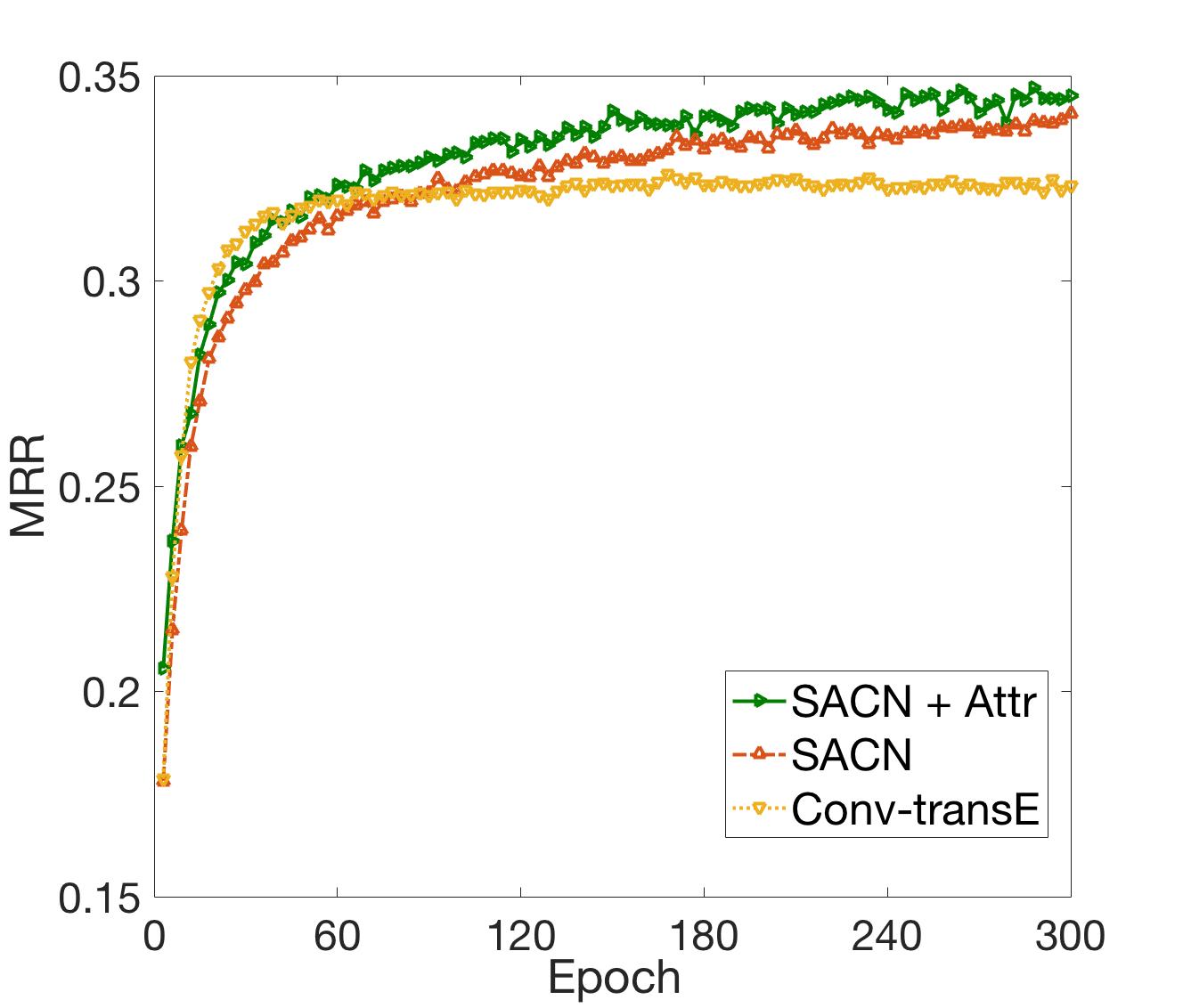}   \\
  \end{tabular}
  \caption{The convergence study of {\it SACN}, {\it Conv-TransE} models in FB15k-237 and {\it SACN} in FB15k-237-Attr ({\it SACN + Attr}) using the validation set. Due to the page limitation, only the results of Hits@1 and MRR are reported here.}
  \label{fig:fb15k-237-analysis}
\end{figure}

\begin{table}
    \caption{Kernel size analysis for FB15k-237 and FB15k-237-Attr datasets. ``{\it  SACN+Attr}" means the {\it SACN} using FB15k-237-Attr dataset.}
    \bigskip
    \label{tab:table4}
    \tabcolsep=0.155cm
    \centering
    \begin{tabular}{c|c|c|c|c|c}
        \hline
        & & \multicolumn{4}{c}{\textbf{FB15k-237}} \\
        \cline{3-6}
        & & \multicolumn{3}{c|}{Hits} & \multicolumn{1}{c}{}  \\
        \cline{3-6}
        Model & Kernel Size & @10 & @3 & @1 & MRR  \\
        \hline
        \hline
        Conv-TransE & 2 $\times$ 1 & 0.504 & 0.357 & 0.234 & 0.324 \\
        \hline
        Conv-TransE & 2 $\times$ 3  & 0.513  & 0.365 & 0.240 & 0.331  \\
        \hline
        Conv-TransE & 2 $\times$ 5 & 0.512 & 0.361 & 0.239 & 0.329  \\
        \hline
        \hline
        SACN & 2 $\times$ 1  & 0.527 & 0.379 & 0.255 & 0.345  \\
        \hline
        SACN & 2 $\times$ 3 & 0.536 & 0.384 & 0.260 & 0.351  \\
        \hline
        SACN & 2 $\times$ 5  & 0.536 & 0.385 & 0.261 & 0.352  \\
        \hline
        \hline
        SACN+Attr & 2 $\times$ 1 & 0.535 & 0.384 & 0.260 & 0.351\\
        \hline
        SACN+Attr & 2 $\times$ 3 & 0.543 & 0.394 & 0.268 & 0.360  \\
        \hline
        SACN+Attr & 2 $\times$ 5 & \textbf{0.547} & \textbf{0.396} & \textbf{0.268} & \textbf{0.360} \\
        \hline

    \end{tabular}
\end{table}

\subsubsection{Convergence Analysis}
Figure \ref{fig:fb15k-237-analysis} shows the convergence of the three models. 
We can see that the {\it SACN} (the red line) is always better than {\it Conv-TransE} (the yellow line) after several epochs. And the performance of {\it SACN} keeps increasing after around 120 epochs. However, the {\it Conv-TransE} has achieved the best performance after around 120 epochs. The gap between these two models proves the usefulness of structural information.
When using the FB15k-237-Attr dataset, the performance of ``{\it SACN + Attr}" is better than ``{\it SACN}" model. 

\subsubsection{Kernel Size Analysis}

In Table \ref{tab:table4}, different kernel sizes are examined in our models. The kernel of ``$2 \times 1$" means the knowledge or information translating between one attribute of entity vector and the corresponding attribute of relation vector. 
If we increase the kernel size to ``$2 \times k$" where $k = \{3,5\}$, the information is translated between a combination of $s$ attributes in entity vector and a combination of $k$ attributes in relation vector. The larger view to collect attribute information can help to increase the performance as shown in Table \ref{tab:table4}. All the values of Hits@1, Hits@3, Hits@10 and MRR can be improved by increasing the kernel size in the FB15k-237 and FB15k-237-Attr datasets. However, the optimal kernel size may be task dependent.

\begin{table}
    \caption{Node indegree study using FB15k-237 dataset.}
    \bigskip
    \label{tab:table5}
    \tabcolsep=0.25cm
    \centering
    \begin{tabular}{c|c|c|c|c}
        \hline
         & \multicolumn{2}{c|}{\textbf{Conv-TransE}}& \multicolumn{2}{c}{\textbf{SACN}} \\
        \cline{2-5}
         & \multicolumn{2}{c|}{Average Hits} & \multicolumn{2}{c}{Average Hits}  \\
        \cline{2-5}
         Indegree Scope & @10 & @3 & @10 & @3  \\
        \hline
        \hline
         [0,100] & 0.192 & 0.125 & 0.195 & 0.134 \\
        \hline
         [100,200] & 0.441 & 0.245 & 0.441 & 0.253  \\
        \hline
         [200,300] & 0.696 & 0.446 & 0.705 & 0.429  \\
        \hline
        [300,400]  & 0.829 & 0.558 &0.806 & 0.577  \\
        \hline
        [400,500] & 0.894 & 0.661 & 0.868 & 0.663  \\
        \hline
        [500,1000]  & 0.918 & 0.767 & 0.891 & 0.695  \\
        \hline
        [1000, maximum]& 0.992 & 0.941 & 0.981 & 0.922\\
        \hline
    \end{tabular}
\end{table}

\subsubsection{Node Indegree Analysis}
The indegree of the node in knowledge graph is the number of edges connected to the node. The node with larger degree means it have more neighboring nodes, and this kind of nodes can receive more information from neighboring nodes than other nodes with smaller degree. As shown in Table \ref{tab:table5}, we present the results for different sets of nodes with different indegree scopes. The average Hits@10 and Hits@3 scores are calculated. Along the increasing of indegree scope, the average value of Hits@10 and Hits@3 will be increased. 
First for a node with small indegree, it benefits from aggregation of neighbor information from the WGCN layers of {\it SACN}. Its embedding can be estimated robustly. Second for a node with high indegree, it means that a lot more information is aggregated through GCN, and the estimation of its embedding is substantially smoothed among neighbors. Thus the embedding learned from {\it SACN} is worse than that from {\it Conv-TransE}. One solution to this problem would be neighbor selection as in \cite{ying2018kdd}.

\section{Conclusion and Future Work}

We have introduced an end-to-end structure-aware convolutional network ({\it SACN}).
The encoding network is a weighted graph convolutional network, utilizing knowledge graph connectivity structure, node attributes and relation types. {\it WGCN} with learnable weights has the benefit of collecting adaptive amount of information from neighboring graph nodes. In addition, the entity attributes are added as the nodes in the network so that attributes are transformed into knowledge structure information, which is easily integrated into the node embedding. 
The scoring network of {\it SACN} is a convolutional neural model, called {\it Conv-TransE}. It uses a convolutional network to model the relationship as the translation operation and capture the translational characteristic between entities and relations. We also prove that {\it Conv-TransE} alone has already achieved the state of the art performance. The performance of {\it SACN} achieves overall about 10\% improvement than the state of the art such as {\it ConvE}.

In the future, we would like to incorporate the neighbor selection idea into our training framework, such as, importance pooling in \cite{ying2018kdd} which takes into account the importance of neighbors when aggregating
the vector representations of neighbors. We would also like to extend our model to be scalable with larger knowledge graphs encouraged by the results in \cite{ying2018kdd}.

\section{Acknowledgements}
This work was partially supported by NSF grants CCF-1514357 and IIS-1718738, as well as NIH grants R01DA037349 and K02DA043063 to Jinbo Bi.

\bibliographystyle{aaai}
\bibliography{Shang-bibliography}

\end{document}